# Semantics for Probabilistic Inference

Henry E. Kyburg, Jr.
Departments of Philosophy and Computer Science
University of Rochester
Rochester, N.Y. 14627

## Abstract

A number of writers (Joseph Halpern and Fahiem Bacchus among them) have offered semantics for formal languages in which inferences concerning probabilities can be made. Our concern is different. This paper provides a formalization of non-monotonic inferences in which the conclusion is supported only to a certain degree. Such inferences are clearly 'invalid' since they must allow the falsity of a conclusion even when the premises are true. Nevertheless, such inferences can be characterized both syntactically and semantically. The 'premises' of probabilistic arguments are sets of statements (as in a database or knowledge base); the conclusions categorical statements in the language. We provide standards for both this form of inference, for which high probability is required, and for an inference in which the conclusion is qualified by an intermediate interval of support.

## 1 INTRODUCTION

Probabilistic reasoning, and probabilistic argument, as we are concerned with them here, are reasoning and argument in which the object is to establish the credibility, or acceptability, or probability, of a conclusion on the basis of an argument from premises that to *not* entail that conclusion. Other terms for the process: inductive reasoning, scientific reasoning, non-monotonic reasoning. What we seek to characterize is that form of argument that will lead to conclusions that may, on the basis of new evidence, be withdrawn.

What is explicitly excluded from probabilistic reasoning, in the sense under discussion, is reasoning from one probability statement to another. Genesereth and Nilsson (1987), for example, offer as an example of their "probabilistic logic" the way in which constraints on the probability of Q can be established on the basis of probabilities for P and for $P \Rightarrow Q$. This is not what we are concerned with. We are concerned with the argument from a set of sentences $\Gamma$ to a particular sentence $\phi$, where that argument is not deductive: that is, the case in which $\Gamma$ does not *entail* $\phi$. Arguments of this sort have been considered at length in philosophy, but no consensus regarding their treatment has emerged.

We will offer below a logic governing (or representing) such inferences, together with a semantics that will provide some motivation for the adoption of that logic.

It should be noted that inference of the sort in question is *not* truth preserving. Thus we do not want to apply conventional semantics directly: we are not concerned with the characterization of some form of truth preserving inference. (Note that inference within the framework of the probability calculus (from $P(A) = p$ and $P(B) = q$ to infer $P(A \vee B) \leq p + q$) *is* truth preserving.) We are concerned with inference from $\Gamma$ to $\phi$, where $\Gamma$ does not *entail* $\phi$, but in some sense *justifies* it.

One argument undermining this effort is the argument that such inference is "subjective." We will take for granted, but also attempt to illustrate, that inconclusive arguments can be just as objective as conclusive arguments, and thus that inductive, probabilistic, uncertain argument as much admits of characterization in semantic terms as does deductive argument.

## 2. DEPARTURES FROM CONVENTIONAL ANALYSIS

There are a number of ways in which we must diverge from conventional logical analysis.

First, the most obvious difference is that our inference procedures must be non-monotonic. If a body of evidence $\Gamma$ does not entail that $\phi$, then clearly further evidence could undermine whatever support $\Gamma$ does lend to $\phi$.

Second, this form of inference is not truth-preserving. It is quite possible that all the sentences in $\Gamma$ should be true, while $\phi$ should be false.

Third, we should not expect that this kind of inference should be "consistent" in the sense that the collection of conclusions supported should have a model: we have in mind not only the lottery paradox, but the theory of measurement, in which we should accept what is overwhelmingly probable in any particular case, but at the same time be sure that in the long run, we will accept something that is false.

Fourth, while logic is impotent in the face of inconsistent premises, we would hope that inductive inference could tolerate inconsistent inputs.

Finally, we should expect that the difference between inputs or premises consisting of a collection of sentences and that consisting of a single (conjunctive) sentence might be relevant.

## 3  IMPLICATION AND INDUCTIVE VALIDITY

Conventional wisdom regarding inference is that *A implies B* is true just in case $B$ is true in every model in which $A$ is true. More generally, the set of sentences $\Gamma$ implies $\phi$ just in case $\phi$ is true in every model in which the sentences of $\Gamma$ are true.

A natural generalization of this wisdom might be to demand that an *inductive* inference is valid in case *almost all* of the models in which $\Gamma$ is true are models in which $\phi$ is true. Even better, if we wish to allow for data that is not strictly consistent: we might demand that $\phi$ be true in *almost all* of the models in which *any* consistent subset of $\Gamma$ is true. Explicitly, we might say that $\phi$ is a valid *inductive* consequence of $\Gamma$ just in case $\phi$ is true in almost all of the models corresponding to each of the maximal consistent subsets $\Gamma'$ of $\Gamma$. Even this won't quite do. So we will say that an inductive inference is valid just in case the conclusion is true in almost all of the models of each set of models *picked out* by a selection function that, for every maximal consistent subset of $\Gamma$, picks out a set of models. We will characterize this function as we go along.

## 4  SYNTAX

We begin with a standard first order language. We allow for names, functions, predicates of any -arity. The proof-theoretic relation we are concerned with is $\Gamma \vDash_{(p,q)} \phi$, which says that $\phi$ is 'implied' to the degree $(p,q)$ by the premises (evidence, body of knowledge) $\Gamma$. The "implies" relation will be represented by a model theoretic relation as just described: $\phi$ must be true in a fraction of the models that is at least $p$, and at most $q$ among each of the sets of models picked out. For inductive logic we may settle for "at least $p$," but for generality we may well be advised to keep track of the upper limit as well.

We adopt the technique used by both Halpern (1990) and Bacchus (1991) of using a two-sorted language, with separate variables for objects and for 'field' terms, intended to be interpreted as real numbers. We include the constants 0 and 1, and the standard operations and relations for real numbers. We include a countable set T of canonical terms representing real numbers between 0 and 1, inclusive (for example decimals). The only syntactic novelty in the language is the inclusion of a primitive relation that connects formulas and real numbers, '$\%(\varphi,\phi,p,q)$,' which says roughly that the proportion of objects satisfying the open formula $\phi$, that also satisfy the open formula $\varphi$ is between $p$ and $q$, inclusive.

## 5  SEMANTICS

A model of our language is exactly the usual sort of thing, a pair $\langle D, \vartheta \rangle$ where $D$ is a domain of individuals, which we constrain to be finite, and $\vartheta$ is an interpretation function. (The field terms take their values in the domain of the reals, and the relations and functions applicable to field terms have their standard interpretations.) The interpretation function maps individual constants onto members of $D$, predicates onto subsets of $D$, $n$-termed relations onto $D^n$, function terms onto functions from $D^n$ to $D$, etc. The only oddity, for such a rich language, is that we shall take $D$ to be finite (though we need not give its cardinality an upper bound).

Truth is defined relative to a model and a variable assignment, $(M, v)$. The only novelty here is the procedure for assigning truth to the special statistical statements of the form "$\%(\varphi,\phi,p,q)$."

The *satisfaction set* of an open formula $\phi$, relative to a model $M = \langle D, \vartheta \rangle$, is the subset of $D^n$ which $\vartheta$ makes $\phi$ true, where $n$ is the number of variables free in $\phi$. In particular, if $\phi$ is a formula containing $n$ free variables $\langle x_1, x_2, ... x_n \rangle$, then the satisfaction set of $\phi$ is the set of $n$-tuples of $D$ such that

$$\left(M, v^{a_1,...a_n}_{x_1,...x_n}\right) \vDash \phi$$

If $\phi$ is a predicate followed by $n$ variables, the satisfaction set of $\phi$ is just its interpretation in $M$. If $\phi$ is a predicate followed by $k$ terms, the satisfaction set of $\phi$ is the set of objects $\langle a_1, a_2, ...a_n \rangle \in D^n$ such that the assignment of those objects to variables of $\phi$ render $\phi$ true. (Thus $S(B(x,a))$ could be the set of individuals in $D$ who are smart ('$S$') brothers ('$B$') of $a$.) If $\phi$ and $\varphi$ contain exactly the same free variables, the satisfaction set of '$\varphi \wedge \phi$' is the intersection of the satisfaction sets of $\varphi$ and of $\phi$. If $\phi$ contains $n$ free variables, the satisfaction set of '$\neg \phi$' is the complement of the satisfaction set of $\phi$ in $D^n$. Since the satisfaction set of an open formula $\phi$ depends only on the domain and the interpretation function $\vartheta$, we will denote it by $\vartheta(\phi)$.

We say that "$\%(\varphi,\phi,p,q)$," is *true* in $(M, v)$, if and only if

(i)     $p$ and $q$ are field variables or members of T, the set of canonical terms,

(ii)    $\varphi$ and $\phi$ are open formulas, and the ratio of the cardinality of the satisfaction set assigned by $\vartheta$ to '$\varphi \wedge \phi$' to the cardinality of the satisfaction set assigned by $\vartheta$ to $\phi$ is between the real number assigned by $(M, v)$ to $p$ and the real number assigned by $(M, v)$ to q. (If no objects are assigned by $\vartheta$ to $\phi$, then the formula is to be regarded as true if $p = 0$, and false otherwise: $\%(\varphi,\phi,p,q) \equiv |\vartheta(\phi)|*p \le |\vartheta(j \ \ddot{Y} \ f)| \le |\vartheta(f)|*q.$[1])

"%" is thus a general variable binding operator, and represents something like relative frequency in the (finite)

---

[1] We could achieve greater elegance and generality by incorporating the features of Popper functions in our interpretation. The costs, relative to the benefits, seem excessive.

model under consideration. If $p$ or $q$ are field variables, then the truth of a statistical statement in which they occur depends on the variable assignment; otherwise (since otherwise they are real number terms in canonical form) they do not.

Examples: "%(R(x,y),Q(x,y),p,q)" is true in a model just in case in that model the proportion of pairs satisfying "R(x,y)&Q(x,y)", among the pairs satisfying "Q(x,y)" is in the interval [p,q].

"%(R(x,y),Q(w,y),p,q)" is true in a model just in case in that model the proportion triples satisfying "R(x,y)&Q(w,y)" among the pairs satisfying "Q(w,y)" is in [p,q]. (There can't be any!)

"%(R(x,y),Q(w,v),p,q)" is true in a model just in case in that model the proportion of quadruples satisfying "R(x,y)&Q(w,v)" among the pairs satisfying "Q(w,v)" is in [p,q], namely, none.

"%(R(x,y),Q(F(x),G(x,y)),p,q)" is true in a model just in case in that model the proportion of pairs <x,y> such that the F of x and the G of the pair <x,y> stand in the relation Q, are also such that they (x and y) satisfy R, lies between p and q.

We now have truth defined for all the sentences of our formal language relative to models and variable assignments having a finite object domain $D$.

## 6 PROBABILISTIC SOUNDNESS

While soundness for valid deductive inference consists in the fact that the conclusion is true in *every* model in which the premises are true, probabilistic soundness must meet a weaker condition. As pointed out above, the conditions for inductive validity are not quite a simple weakening, but we will start by requiring that a valid inductive (or uncertain) inference of force $[p,q]$ be characterized by the fact that of the models in which the premises are true, between $p$ and $q$ are such that the conclusion is true.

This doesn't seem much like inference, but if we think of $p$ as a number close to 1, and disregard $q$, validity in this sense seems like an appropriate standard of inductive acceptance. We will, however, continue to treat the question generally, i.e., define validity relative to an interval $[p,q]$, where $p$ and $q$ are field terms, and not variables.

The first thing that falls out of this approach is that statements *known* to be equivalent will receive the same (interval) degree of support:

T-1  If '$\phi \equiv \varphi$' is in $\Gamma$, then $\Gamma \vDash_{(p,q)} \phi$ if and only if $\Gamma \vDash_{(p,q)} \varphi$.

If the biconditional is in the set of premises, any model that makes the premises true will either make both $\varphi$ and $\phi$ true, or neither true. This is so, even if there is a selection function that picks a subset of this set of models. Thus the proportion of models that make the premises true that also make the conclusion will be the same in either case.

## 7 PREMISES

Before we go further, we must give some thought to $\Gamma$. What we have in mind is a set of sentences representing a body of knowledge or a database. Typically it will be large. Typically it will have both sentences of the form '$P(a)$' attributing a property to an object, and sentences of the form '%($\varphi,\phi,p,q$)' constraining the proportion of objects having one property, among those that have another property. It may also have sentences of other forms, but it is sentences of these two forms, together with biconditionals, that will mainly concern us. We should not think of $\Gamma$ as a deductively closed set of sentences. It will be better to think of $\Gamma$ as a finite set of sentences given explicitly; for example, as axioms. For the time being we will assume that this set of sentences is simultaneously satisfiable by at least one model, but later we will lift that restriction to show that probabilistic inference does not (always) break down in the face of inconsistency in the database.

Although $\Gamma$ should be thought of as finite, and not deductively closed, we will still assume that any logically or mathematically valid formula can be added to it. What this means, of course, is that when a statement is shown to be valid, we are justified in adding it to $\Gamma$.

## 8 NUMBERS OF MODELS

Suppose that $\Gamma$ consists of the two sentences, "%($A(x),B(x),p,q$)" and "$B(a)$." Then the number of models in which "%($A(x),B(x),p,q$)" is true can be calculated for $D$ of given size; it is merely a combinatorial problem to calculate the number of ways in which $|D|$ objects can be assigned to four boxes, $AB$, $A\neg B$, $\neg AB$, and $\neg A\neg B$ in such a way that $|AB|/|B|$ lies in $[p,q]$. In that set of models, we pick out the subset in which "$B(a)$" is true. But without even knowing the number of models satisfying the premises, we can be quite sure than between $p$ and $q$ of them also satisfy "$A(a)$," since the term $a$ may denote any of the objects in the box $B$. It follows that:

T-2  {"%($A(x),B(x),p,q$)", "$B(a)$"} $\vDash_{(p,q)}$ "$A(a)$."

If $p$ is large enough, say greater than some selected ratio $1 - \delta$, then we may accept "$A(a)$" on the basis of the *total* evidence {"%($A(x),B(x),p,q$)","$B(a)$"}. Clearly, if we know more than this, the inference may be undermined. But so may the assignment of probability in T-2. This form of argument is clearly non-monotonic.

This is not $\varepsilon$-semantics[2]: we intend $\delta$ to be a fixed real number, chosen according to the demands of the context.[3]

---

[2] Geffer and Pearl (1990), Pearl (1988).

[3] Some suggestions concerning how the level of acceptance might be chosen were given in (Kyburg, 1988).

We do not demand that the chance of a mistake be reduced to an arbitrarily low level, but only that it be reduced to less than $\delta$.

## 9  SPECIFICITY

A natural constraint on uncertain or probabilistic inference, well known since Reichenbach[4], is that given a choice between two reference classes for a probability, if we know one is a subclass of the other it is a better guide to belief. In the theory of evidential probability (Kyburg 1991a), the included class is also to be preferred. In other contexts 'specificity' is also deemed appropriate (Touretzky 1986).

Suppose that we add "$B'(a)$," "$\forall x(B'(x) \to B(x))$," and "$\%(A(x),B'(x),p',q')$" to our previous database, where $p <p'$ & $q < q'$ or $p >p'$ & $q > q'$. Then we should expect that the proportion of models in which the premises are true would be between $p'$ and $q'$. Indeed, this is so, for the set of models in which "$B'(a)$" is satisfied will be constrained to be a *subset* of the set of models in which "$B(a)$" is satisfied, and in that subset, by the same argument as before, the proportion of models in which "$A(a)$" is satisfied will be between $p'$ and $q'$.

T-3   If $p <p'$ & $q < q'$ or $p >p'$ & $q > q'$, then
{"$\%(A(x),B(x),p,q)$","$B(a)$" "$B'(a)$," "$\forall x(B(x) \mathrel{\text{Æ}} B'(x))$," "$\%(A(x),B'(x),p',q')$" } $\models_{(p,q)}$ "$A(a)$."

As indicated in Kyburg (1991b), 'specificity' is only one among three quite different circumstances under which one statistical inference structure *dominates* another. The other two (and, we claim, there are only two others) are the structure which takes account of known prior distributions (the 'Bayesian' relation) and the structure which corresponds to having more data (the 'subsample' relation). We will explore the model-theoretic explanation for these forms of domination in another place.

## 10  STRENGTH

One of the more controversial features of evidential probability is the 'strength principle'. According to this principle, the fact that I know an applicant for insurance is a Mormon does not prevent a broader reference class from being appropriate for determining his premium, unless I also know that the relevant death rate is different among Mormons. To know that the death rate is different is to be construed as in the preceding theorem, that is, as knowing $p <p'$ & $q < q'$ or $p >p'$ & $q > q'$. (Since $p$, $q$, etc., are field terms and not variables, to know this requires only the comparison of fractions in canonical form.)

Suppose that what we know is that $p' \le p$ & $q \le q'$. Then we know both that in the set of models in which "$\%(A(x),B(x),p,q)$" and "$B(a)$" are true "$A(a)$" is true in between $p$ and $q$ of them, and also that in the set of models in which "$\%(A(x),B(x),p,q)$","$B(a)$" "$B'(a)$", "$\forall x(B(x) \to B'(x))$" and "$\%(A(x),B'(x),p',q')$" are all true, between $p'$ and $q'$ are models in which "$A(a)$" is also true. But we already knew that. Adding the new sentences doesn't *change* the force of the probabilistic argument so much as it weakens it. We can capture this insight semantically by saying that the combined evidence picks out a subset of models satisfying not only the constraint embodied in "$\forall x(B(x) \to B'(x))$", and the constraint embodied in "$\%(A(x),B'(x),p',q')$", but also, in view of the lack of conflict between "$\%(A(x),B(x),p,q)$" and "$\%(A(x),B'(x),p',q')$", having a proportion $[\vartheta(A) \cap \vartheta(B')]/\vartheta(B')$ that lies between $p$ and $q$, rather than merely between $p'$ and $q'$. Clearly this is always possible.[5] This is an operation of the selection function: the premises ($\Gamma$) pick out a set of models in which the premises are all true; the selection principle picks out a further subset, corresponding to the fact that the additional constraint (to models in which "$B'(a)$" is true) does not add to our evidence concerning "$A(a)$".

It is important to note that we are *not* assuming that "$\%(A(x),B'(x),p,q)$" should be added to our database, but only that the measure of support or degree of implication of "$A(a)$" given by our database is determined by the set of models in which that statement *would* be true.

T-4   If $p' \le p$ & $q \le q'$, then
{"$\%(A(x),B(x),p,q)$","$B(a)$" "$B'(a)$," "$\forall x(B(x) \to B'(x))$," "$\%(A(x),B'(x),p',q')$" } $\models_{(p,q)}$ "$A(a)$."

## 11  CONFLICTING EVIDENCE

Suppose that we know that $a$ has both the properties $B$ and $B'$, and that the statistics we have differ, as in T-3, but that we do not know that $B$ is included in $B'$, or vice versa. The models of {"$\%(A(x),B(x),p,q)$", "$B(a)$", "$B'(a)$", "$\%(A(x),B'(x),p',q')$", $B'(a)$} differ in the proportion of $A$'s among $B$'s and among $B'$'s. We have conflicting evidence. The selection function picks out a set of models concerning which the statistical premises of $\Gamma$ agree: namely, those in which both $[\vartheta(A) \cap \vartheta(B)]/\vartheta(B)$ and $[\vartheta(A) \cap \vartheta(B')]/\vartheta(B')$ are bounded only by $\min(p,p')$ and $\max(q,q')$. This is a *larger* set of models, rather than a smaller one, but in this set there is no conflict between $B$ and $B'$.

T-5   If $p <p'$ & $q < q'$ or $p >p'$ & $q > q'$, then
{"$\%(A(x),B(x),p,q)$","$B(a)$","$B'(a)$", "$\%(A(x),B'(x),p',q')$", $B'(x)$] $\models_{(\min(p,p'),\max(q,q'))}$ "$A(a)$."

## 12  THE GENERAL CASE

---

[4] Hans Reichenbach,: *The Theory of Probability*, University of California Press, Berkeley and Los Angeles, 1949. (German ed. 1934.)

[5] The same idea will be found in Bacchus (1990), employed qualitatively to deal with 'overly specific reference classes' (p. 162): if we have no information, then we may assume that the measure in the subset is the same as that in the superset. We are here not making this (generally unwarranted) assumption.

Note that each of these theorems, we are taking the listed members of $\Gamma$ to be the only members of $\Gamma$. None of these theorems thus reflect more than ideally impoverished bodies of knowledge that cannot be taken very seriously. In any realistic situation, we must take account of far richer background knowledge. A variety of syntactical work has been done (Kyburg 1961, Kyburg and Murtezaoglu 1991, Loui 1988) with the object of providing rules and algorithms for implementing this notion of inductive support, where the contents of the premise set is plausibly rich.

There appears to be no reason not to suppose that the semantics for the richer cases cannot be worked through. In general there are a number of possible ways of deriving a probability for a statement, some of which involve stronger statistical statements than others, some of which involve differing statistical statements, and various relations may be known among the potential references classes. By T–1, all these differences must be resolved.

Additions to the selection function are required to deal with two further extremely important cases in which G embodies conflicting evidence: the case in which taking account of *known* prior distributions is important, and the general case of statistical inference. These have been alluded to elsewhere (Kyburg, 1991b), and will not further occupy us here.

## 13   INCONSISTENCY

Let us now turn to the case in which $\Gamma$ is inconsistent, i.e., has no model. How can inductive inference be possible? A maximal consistent subset of the finite set $\Gamma$ is a set of sentences, each of which belongs to $\Gamma$, and which is such that if any other sentence of $\Gamma$ is added to it, inconsistency results. Let us suppose we have $\Gamma' \vDash_{(p,q)} \phi$ for every maximal consistent subset $\Gamma'$ of $\Gamma$. Then we should have:

T–6    If $\Gamma' \vDash_{(p,q)} \phi$ for every maximal consistent subset $\Gamma'$ of $\Gamma$, then $\Gamma \vDash_{(p,q)} \phi$.

A familiar example in which this would make sense is the following: let $\Gamma$ contain a description of the million ticket lottery, and, for every $i$ from 1 to a million, the sentence "$\neg W(t_i)$". Let it also contain "$\%(H,T,.49,.51)$" and "$T(a)$", where $T$ is the predicate "-is a toss of a coin" and $H$ is the predicate "-lands heads up". If we assume that description of the lottery says that there is one winner, then there is no model of this set of sentences. Intuitively there is no reason why we should not be able to infer, with confidence [.49,.51], that "$H(a)$". And indeed, T–6 leads to this result.

On the other hand, we cannot infer anything interesting about "$\neg W(t_1)$". There are models (those in which $\vartheta(t\ 1)$ is a member of $\vartheta(W)$) in which "$\neg W(t_1)$" is false, and in *all* of those models "$W(t_1)$" is true, so that $\Gamma \vDash_{(1,1)}$ "$W(t_1)$". There are also a lot of models in which "$\neg W(t_1)$" is true, so that $\Gamma \vDash_{(0,0)}$ "$W(t_1)$". Thus, relative to the database $\Gamma$, there is no probability that ticket $t_1$ will lose, and in particular no high probability that it will lose.

We may state this as T–7:

T–7    It is possible that $\phi \in \Gamma$, while $\Gamma \vDash_{(p,q)} \phi$ is false for all $p$ and $q$, and thus that f may not be inductively inferred from $\Gamma$ at any level of confidence.

On the other hand, if $\Gamma$ is consistent, and $\phi \in \Gamma$, then $\Gamma \vDash_{(1,1)} \phi$, as may be seen thus: if $\phi \in \Gamma$ and $\Gamma$ is consistent, then every sentence in $\Gamma$ holds in every model of $\Gamma$ — we only turn to maximal consistent subsets of $\Gamma$ when $\Gamma$ has no model. But then $\phi$ is true in every model of $\Gamma$ and therefore in every selected model of $\Gamma$. Thus,

T–8    If $\Gamma$ has a model and $\phi \in \Gamma$, then $\Gamma \vDash_{(1,1)} \phi$.

This insulation from the effects of inconsistency makes probabilistic logic as construed here seem an interesting alternative to ordinary first order logic. Settle on a value for $\delta$, and accept as inductively justified any statement whose lower probability, relative to the database $\Gamma$, exceeds $1 - \delta$. The resulting set of inductively justified sentences will include the logical and mathematical truths in $\Gamma$. It may be inconsistent (for example if $\Gamma$ contains a description of the lottery), but it will not *inherit* inconsistencies from $\Gamma$. We conjecture that there may be a gain in computational efficiency in taking probabilistic logic as the basic logic for making inferences from $\Gamma$.

## 14   RELATION TO OTHER WORK

Charles Morgan (1991) has examined an approach to founding various logics on probabilistic principles. He describes a variety of logics in probabilistic terms (and even provides a handy recipe for generating them!), but since he takes a conditional probability of 1 (relative to a condition whose probability is non-0) to correspond to validity, it is clear that his probabilistic logic differs from the one offered here. The contrast can be rendered stark by considering four principles Morgan lays down as basic to his enterprise:[6]

R.1.   If $A \in \Gamma$ then $\Gamma \vdash A$.

R.2.   If $\Gamma \vdash A$ then for some finite subset $\Delta$ of $\Gamma$, $\Delta \vdash A$.

R.3.   If $\Gamma \cup \{A\} \vdash B$ and $\Gamma \vdash A$, then $\Gamma \vdash B$.

R.4.   If $\Gamma \vdash A$ then $\Gamma \cup \Delta \vdash A$.

If the syntax developed elsewhere is adequate to the semantics given here, we may interpret the turnstile as follows: $\Gamma \vdash A$ just in case $\Gamma \vDash_{(1-\delta,q)} A$, for given $\delta$ and arbitrary $\theta$. We have just seen that R.1. does not hold for our notion of inductive or probabilistic inference, unless $\Gamma$ is strictly consistent. R.2. holds, since our models are all finite. R.3. may fail to hold for an interesting reason.

---

[6] p. 98.

Let $\Gamma$ contain "$T(a)$" and a large amount of statistical data concerning $T$ and $B$. Let $A$ be "$\%(T(x),B(x),1-\delta,1)$". By T-2 we see that $\Gamma \cup \{A\} \models_{(1-\delta,1)} "T(a)"$. It is possible to show, assuming suitable statistical data is in $\Gamma$, that "$\%(T(x),B(x),1-d,1)$" may be a conclusion that can be validly (inductively) inferred from $\Gamma$, i.e, we can have $\Gamma \models_{(1-d,1)} "\%(T(x),B(x),1-d,1)"$. And yet, since the statistical statement in question is not (by hypothesis) in $\Gamma$, we will not have $\Gamma \models_{(1-d,1)} "T(a)"$. This illustrates two things: that we must distinguish between *evidence* for a probabilistically inferred conclusion, and the *conclusion* itself (we must not just add the inductively inferred conclusion to $\Gamma$); and the deployment of statistical evidence (in the sense of samples drawn) in uncertain inference must be indirect. Finally, R.4. constitutes a rejection of the very non-monotonicity we seek to model through probabilistic inference; we wouldn't want R.4. to hold.

It might be asked why we relate acceptability (or probability) to the mere number of models in which a statement is true. Why not *weight* the models, as, in effect, both Bacchus and Halpern suggest? The basic answer is that at some point you either introduce a subjective measure, or you introduce some method of counting. Halpern introduces a measure on possible worlds (or models) for the sake of generality; Bacchus introduces a weighting on the objects in the domain to account for the fact that in a Bayesian assignment of probability it may not be the case that every object in $D$ gets the same chance of being selected.[7] But we are not talking of 'selecting' objects from $D$, nor are we seeking generality. We are seeking objectivity, and we therefore would prefer to replace subjective 'weighting' with objective 'counting.' This means that we do not regard individual objects in the domain $D$ as being 'chosen', as though they could be chosen many times, and each might be chosen equally often (or not equally often) in the long run. *If* something is chosen, and that is relevant to our assessments of probability, then $D$ should contain, not objects, but choices of objects, each of which in turn should be counted *once* . Thus the argument in favor of counting as opposed to weighting is that if we have some objective measure, and if it disagrees with the weighting that seems appropriate, we can achieve a match by altering the *objects* that are counted, for example from *kinds* of physical objects to *trials* of a chance set-up.

In the interest of both objectivity and computability, we claim that the semantic condition of at-least-(1-$\delta$)-of-the-models captures what we want to capture for probabilistic or inductive inference.

## 15 CONCLUSION

Insofar as there is a sense to be found for 'probabilistic argument' in ordinary language, it is a form of argument in which the conclusion, while not entailed by the premises, achieves such a high degree of certainty that its acceptance is justified: "Our premises justify the acceptance of S." It is not a form of argument in which the conclusion has the form of a probability statement: "S is, relative to our premises, highly probable."

The idea of calling such argument 'probabilistic' is that it is high probability that warrants the acceptance of conclusions of this sort of argument. High probability, however, has the problem that it leads to the acceptance of inconsistent conclusions, as shown by the lottery paradox (Kyburg, 1961). Furthermore, treatments of acceptance by virtue of high probability have been syntactic in character. What we have attempted here is to give a semantics for probabilistic inference. But if there is a semantic relation that validates probabilistic inference, it is also a relation that characterizes propositional probability generally. We have therefore attempted to sketch a semantics (incomplete) for the relation $\models_{(p,q)}$, of which the relation warranting acceptance at level 1-$\delta$, $\models_{(1-\delta,q)}$, is a special case.

The results of this semantic analysis are that truth in most models (most selected models) provides an intuitive warrant for the syntactic relation of high probability relative to premises as a ground for credence, that truth in most models is a reasonable ground for acceptance (with 'most' defined in terms of 1 - $\delta$), and that the awkwardness of inconsistent databases need not make difficulty for probabilistic inference thus construed. It remains to be shown that the relation of high probability relative to a database is more efficiently computed than the relation of entailment by the database (assuming that it's so!).

### Acknowledgement

Support of the National Science Foundation is gratefully acknowledged.

### References

Genesereth, Michael and Nilsson, Nils (1987). *Logical Foundations of Artificial Intelligence*, Morgan Kaufmann, Los Altos, 1987.

Halpern, Joseph (1990). "An Analysis of First-Order Logics of Probability," *A I Journal* 46, 1990 pp 311-350.

Bacchus, Fahiem (1991). *Representing and Reasoning with Probabilistic Knowledge*, The MIT Press, Cambridge, 1991.

Kyburg, Henry E., Jr. (1991a). "Evidential Probability," *IJCAI-91: Proceedings*, 1991a, pp 1196–1202.

Touretzky, D. S. (1986). *The Mathematics of Inheritance Systems*, Morgan Kaufman, Lost Altos, 1986.

Geffner, Hector and Pearl, Judea (1991). "A Framework for Reasoning with Defaults," in Kyburg, Loui, and Carlson (eds) *Knowledge Representation and Defeasible Reasoning*, Kluwer, Dordrecht 1991, pp. 69-88.

---

[7] In our systems this is accounted for by adopting a more general notion of 'object' which may be construed as a trial of a certain sort.

Kyburg, Henry E. Jr. (1988). "Full Belief." *Theory and Decision* **25**, 1988, 137-162.

Reichenbach, Hans (1949). *The Theory of Probability*, University of California Press, Berkeley and Los Angeles, 1949. (German ed. 1934.)

Kyburg, Henry E., Jr. and Murtezaoglu, Bulent (1991). "A Modification to Evidential Probability," *Uncertainty in Artificial Intelligence, Proceedings*, 1991, pp. 228–231.

Loui, Ronald P. (1986). "Computing Reference Classes," *Proceedings of the 1986 Workshop on Uncertainty in Artificial Intelligence*, (1986) 183-188.

Kyburg, Henry E., Jr. (1961). *Probability and the Logic of Rational Belief*, Wesleyan University Press, Middletown, Ct. 1961

Kyburg, Henry E., Jr.(1991b) "Beyond Specificity," in B. Bouchon-Meunier, R. R. Yager, and L. A. Zadeh (eds), *Uncertainty in Knowledge Bases,* Lecture Notes in Computer Science, Springer Verlag, 1991b, pp.204-212.

Morgan, Charles G. (1991)."Logic, Probability, and Artificial Intelligence," *Computational Intelligence* **7**, 1991, pp. 94-109.